\def\year{2019}\relax

\documentclass[letterpaper]{article} 
\usepackage{aaai19}  
\usepackage{times}  
\usepackage{helvet}  
\usepackage{courier}  
\usepackage{url}  
\usepackage{graphicx}  
\usepackage{amsmath}
\usepackage{amssymb}
\usepackage{amsthm}
\usepackage{multirow}
\usepackage{etoolbox}
\usepackage{mathrsfs}
\usepackage{booktabs}
\usepackage{subcaption}

\DeclareMathOperator*{\argmax}{arg\,max}
\newcommand{\MyMapTemplatePrefix}[4]{\expandafter#1\csname#3#4\endcsname{#2{#4}}}
\newcommand{\MyMapTemplatePrefixNew}[5]{\expandafter#1\csname#4#5\endcsname{#2{#3{#5}}}}
\forcsvlist{\MyMapTemplatePrefix {\def} {\mathbf} {}} {A,B,C,D,E,F,G,H,I,J,K,L,M,N,O,P,Q,R,S,T,U,V,W,X,Y,Z}
\forcsvlist{\MyMapTemplatePrefix {\def} {\mathbf} {}} {a,b,c,d,e,f,g,h,i,j,k,l,m,n,o,p,q,r,s,t,u,v,w,x,y,z,1,0}
\forcsvlist{\MyMapTemplatePrefix {\def} {\widetilde} {wt}} {A,B,C,D,E,F,G,H,I,J,K,L,M,N,O,P,Q,R,S,T,U,V,W,X,Y,Z}
\forcsvlist{\MyMapTemplatePrefix {\def} {\widetilde} {wt}} {a,b,c,d,e,f,g,h,i,j,k,l,m,n,o,p,q,r,s,t,u,v,w,x,y,z} 
\forcsvlist{\MyMapTemplatePrefixNew {\def} {\widetilde}{\mathbf} {tb}} {A,B,C,D,E,F,G,H,I,J,K,L,M,N,O,P,Q,R,S,T,U,V,W,X,Y,Z}
\forcsvlist{\MyMapTemplatePrefixNew {\def} {\widetilde}{\mathbf} {tb}} {a,b,c,d,e,f,g,h,i,j,k,l,m,n,o,p,q,r,s,t,u,v,w,x,y,z}
\forcsvlist{\MyMapTemplatePrefix {\def} {\mathcal}{mc}} {A,B,C,D,E,F,G,H,I,J,K,L,M,N,O,P,Q,R,S,T,U,V,W,X,Y,Z}
\forcsvlist{\MyMapTemplatePrefix {\def} {\mathbb} {mb}} {A,B,C,D,E,F,G,H,I,J,K,L,M,N,O,P,Q,R,S,T,U,V,W,X,Y,Z}

\def\tp{^\intercal} \def\st{\text{s.t.~}}
\def\ie{{i.e.}} \def\etal{{et.al}}
 \def\eg{{e.g.}}
\def\bSigma{\mathbf{\Sigma}} \def\bmu{\mathbf{\mu}}  \def\btheta{\mathbf{\theta}} \def\bTheta{\mathbf{\Theta}}
\def\bsigma{\mathbf{\sigma}}

\newtheorem{defn}{Definition}  
\newtheorem{prop}{Proposition}

\begin{document}
%
\title{Gaussian-Induced Convolution for Graphs}
\author{Jiatao Jiang, Zhen Cui\thanks{Corresponding author}, Chunyan Xu, Jian Yang\\
	PCA Lab, Key Lab of Intelligent Perception and Systems for High-Dimensional Information of Ministry of Education,\\
	and Jiangsu Key Lab of Image and Video Understanding for Social Security,\\
	School of Computer Science and Engineering, Nanjing University of Science and Technology, Nanjing, China\\
	\{jiatao, zhen.cui, cyx, csjyang\}@njust.edu.cn\\
} 

\maketitle

\begin{abstract}
Learning representation on graph plays a crucial role in numerous tasks of pattern recognition. Different from grid-shaped images/videos, on which local convolution kernels can be  lattices, however, graphs are fully coordinate-free on vertices and edges. In this work, we propose a Gaussian-induced convolution (GIC) framework to conduct local convolution filtering on irregular graphs. Specifically, an edge-induced Gaussian mixture model is designed to encode variations of subgraph region by integrating edge information into weighted Gaussian models, each of which implicitly characterizes one component of subgraph variations. In order to coarsen a graph, we derive a vertex-induced Gaussian mixture model to cluster vertices dynamically according to the connection of edges, which is approximately equivalent to the weighted graph cut. We conduct our multi-layer graph convolution network on several public datasets of graph classification. The extensive experiments demonstrate that our GIC is effective and can achieve the state-of-the-art results.
\end{abstract}

\section{Introduction}

\begin{figure*}[t]
	\centering
	\begin{subfigure}{0.15\textwidth}
		\centering
		\includegraphics[width=1\textwidth]{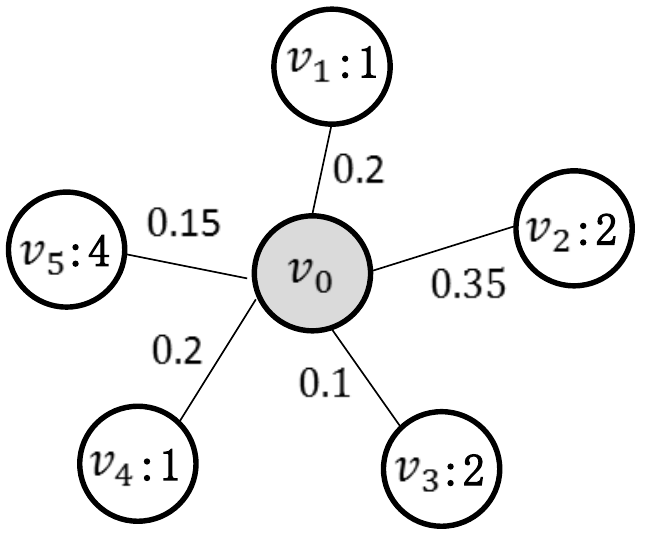}
		\caption{}
		\label{fig:motivation:A}
	\end{subfigure}
	\quad \quad \quad \quad
	\begin{subfigure}{0.17\textwidth}
		\centering
		\includegraphics[width=1\textwidth]{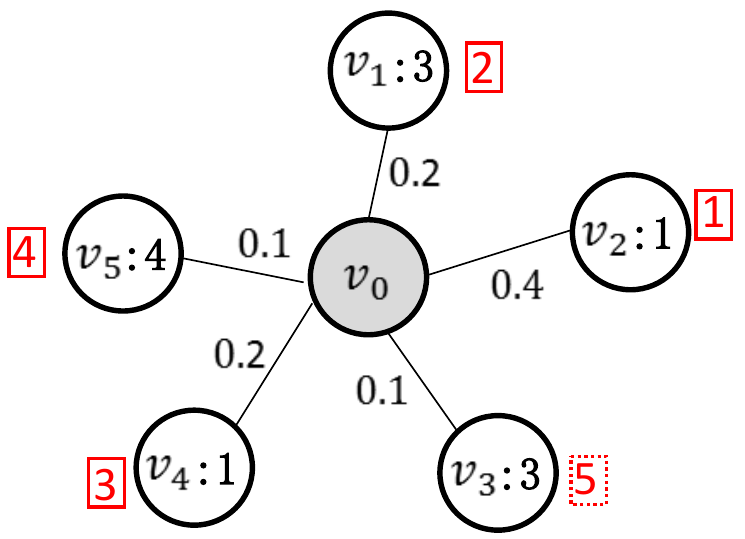}
		\caption{}
		\label{fig:motivation:B}	
	\end{subfigure}
	\quad \quad \quad \quad
	\begin{subfigure}{0.16\textwidth}
		\centering
		\includegraphics[width=1\textwidth]{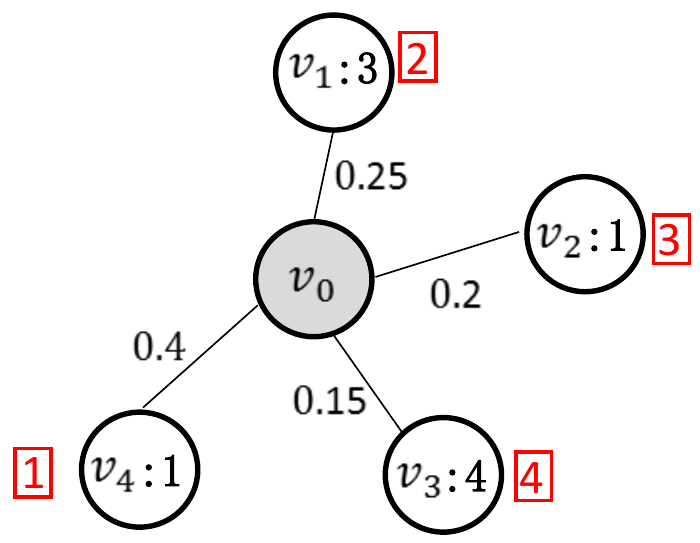}
		\caption{}
		\label{fig:motivation:C}
	\end{subfigure}
	\quad \quad \quad \quad
	\begin{subfigure}{0.15\textwidth}
		\centering
		\includegraphics[width=1\textwidth]{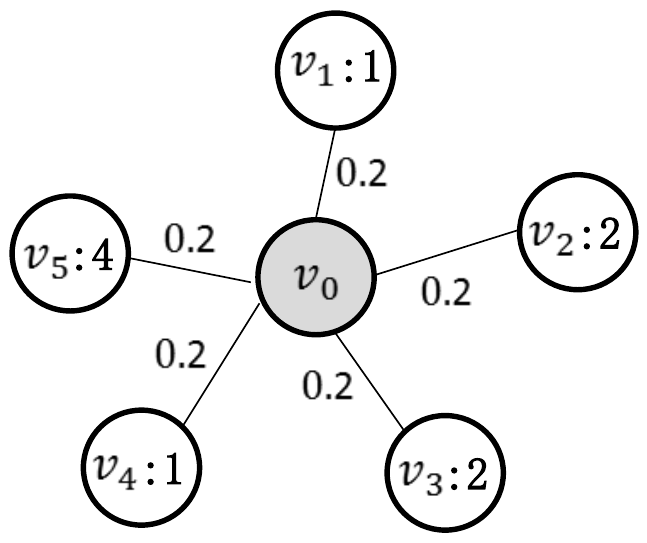}
		\caption{}
		\label{fig:motivation:D}
	\end{subfigure}
	\\ 
	\begin{subfigure}{0.19\textwidth}
		\centering
		\includegraphics[width=1\textwidth]{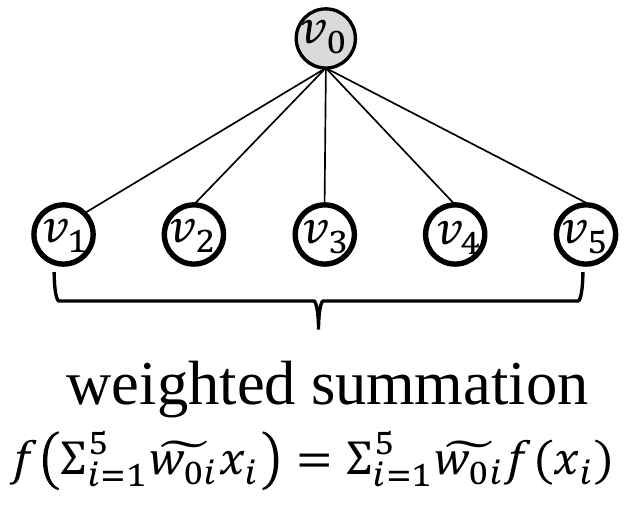}
		\caption{Tied filtering}
		\label{fig:motivation:E}
	\end{subfigure}
	\quad \quad
	\begin{subfigure}{0.23\textwidth}
		\centering
		\includegraphics[width=1\textwidth]{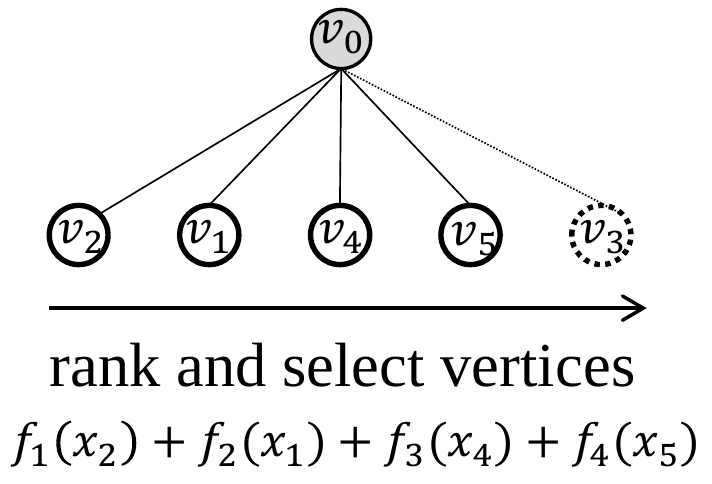}
		\caption{Ranking filtering}
		\label{fig:motivation:F}
	\end{subfigure}
	\quad \quad
	\begin{subfigure}{0.20\textwidth}
		\centering
		\includegraphics[width=1\textwidth]{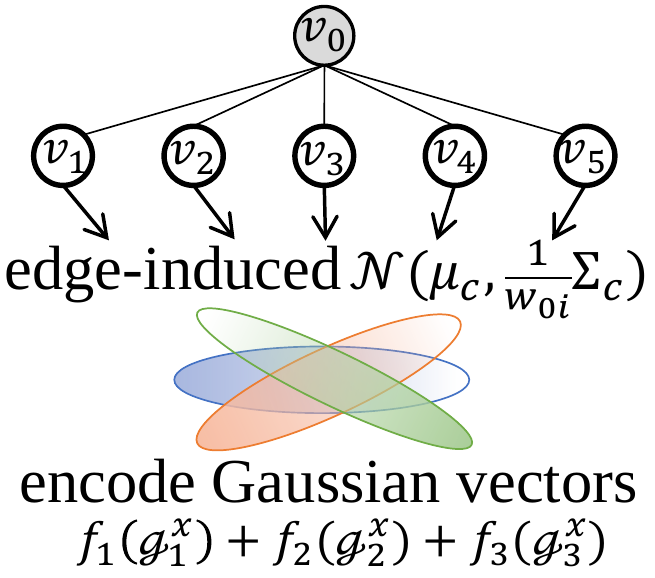}
		\caption{Gaussian-induced filtering}
		\label{fig:motivation:G}
	\end{subfigure}
	\quad \quad
	\begin{subfigure}{0.17\textwidth}
		\centering
		\includegraphics[width=1\textwidth]{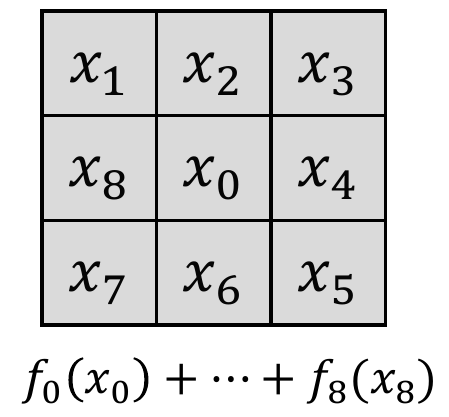}
		\caption{$3\times 3$ regular filtering}
		\label{fig:motivation:H}
	\end{subfigure}
	\caption{Different filters operations on graph vertices. Examples of one-hop subgraphs are given in (a)-(d), where $v_0$ is the reference vertex and each vertex is assigned to a signal. The tied filtering (e) summarizes all neighbor vertices, and generates the same responses to (a) and (b) under the filter $f$, \ie, $f(\sum \widetilde{w_{0i}}x_i)=f(1.9)$, although the two graphs are completely different in structures. The ranking filtering (f) sorts/prunes neighbor vertices and then performs different filtering on them. It might result into the same responses $f_1(1)+f_2(3)+f_3(1)+f_4(4)$ to different graphs such as (b) and (c), where the digits in red boxes denote the ranked indices and the vertex of dashed box in (b) is pruned. Moreover, the vertex ranking is uncertain/non-unique for equal connections in (d).To address these problems, we derive edge-induced GMM to coordinate subgraphs as shown in (g). Each of Gaussian model can be viewed as one variation component (or direction) of subgraph. Like the standard convolution (h), the Gaussian encoding is sensitive to different subgraphs, \eg, (a)-(d) will have different responses. Note $f, f_i$ are linear filters, and the non-linear activation functions are put on their responses.
	}
	\label{fig:motivation}
\end{figure*}

As witnessed by the widespread applications, graph is one of the most successful models to conduct structured and semi-structured data, ranging from text~\cite{defferrard2016convolutional}, bioinformatics~\cite{yanardag2015deep,niepert2016learning,song2018eeg} and social network~\cite{gomez2017dynamics,orsini2017shift} to images/videos~\cite{marino2016more,cui2018context,cui2017spectral}. Among these applications, learning robust representations from structured graphs becomes the main topic. To this end, various methods have come forth in recent years. Graph kernels~\cite{yanardag2015deep} and recurrent neural networks (RNNs)~\cite{scarselli2009graph} are the most representative ones. Graph kernels usually take the classic R-convolution strategy~\cite{haussler1999convolution} to recursively decompose graphs into atomic sub-structures and then define local similarities between them. RNNs based methods sequentially traverse neighbors with tied parameters in depth. With the increase of graph size, graph kernels would suffer diagonal dominance of kernels~\cite{scholkopf2002kernel} while RNNs would have the explosive number of combinatorial paths in the recursive stage.

Recently convolutional neural networks (CNNs)~\cite{lecun2015deep} have achieved breakthrough progresses on representing grid-shaped image/video data. In contrast, graphs are with irregular structures and fully coordinate-free on vertices and edges. The vertices/edges are not strictly ordered, and can not be explicitly matched between two graphs. To generalize the idea of CNNs onto graphs, we need to solve this problem therein that the same responses should be produced for those homomorphic graphs/subgraphs when performing convolutional filtering. To this end, recent graph convolution methods~\cite{defferrard2016convolutional,atwood2016diffusion,hamilton2017inductive} attempted to aggregate neighbor vertices as shown in Fig.~\ref{fig:motivation:E}. This kind of methods actually employ a fuzzy filtering (\ie, a tied/shared filter) on neighbor vertices because only first-order statistics (mean) is used. Two examples are shown in Fig.~\ref{fig:motivation:A} and Fig.~\ref{fig:motivation:B}. Although they have different structures, the responses on them are fully equal. Oppositely, Niepert \etal~\cite{niepert2016learning} ranked neighbor vertices according to weights of edges, and then used different filters on these sorted vertices, as shown in Fig.~\ref{fig:motivation:F}. However, this rigid ranking method will suffer some limitations: i) probably consistent responses to different structures (\eg, Fig.~\ref{fig:motivation:B} and Fig.~\ref{fig:motivation:C}) because weights of edges are out of consideration after ranking; ii) information loss of node pruning for a fixed-size receptive field as shown in Fig.~\ref{fig:motivation:B}; and iii) ranking ambiguity for equal connections as shown in Fig.~\ref{fig:motivation:D}; and iv) ranking sensitivity to (slightly) changes of edge weights/connections.

In this paper we propose a Gaussian-induced graph convolution framework to learn graph representation. For a coordinate-free subgraph region, we design an \textit{edge-induced} Gaussian mixture model (EI-GMM) to implicitly coordinate the vertices therein. Specifically, the edges are used to regularize Gaussian models such that variations of subgraph can be well-encoded. In analogy to the standard convolutional kernel as shown in Fig.~\ref{fig:motivation:H}, EI-GMM can be viewed as a coordinate normalization by projecting variations of subgraph into several Gaussian components. For example, the four subgraphs w.r.t. Fig.~\ref{fig:motivation:A}$\sim$\ref{fig:motivation:D} will have different representations\footnote{Suppose three Gaussian models are $\mcN(0,1), \mcN(0,2)$ and $\mcN(0,3)$, then we can compute the responses on (a)-(d) respectively as $f_1([0.49, -0.93])+f_2([0.17, -0.65])+f_3([0.07, -0.44])$, $f_1([0.35, -0.73])+f_2([0.15, -0.58])+f_3([0.10, -0.64]$, $f_1([0.35, -0.71])+f_2([0.15, -0.39])+f_3([0.10, -0.43])$, $f_1([0.46, -0.99])+f_2([0.18, -0.62])+f_3([0.08, -0.42])$. Please refer to incoming section.} through our Gaussian encoding in Fig.~\ref{fig:motivation:G}. To make the network inference forward, we transform Gaussian components of each subgraph into the gradient space of multivariate Gaussian parameters, instead of employing the sophisticated EM algorithm. Then the filters (or transform functions) are performed on different Gaussian components like latticed kernels on different directions in Fig.~\ref{fig:motivation:H}. Further, we derive a \textit{vertex-induced} Gaussian mixture model (VI-GMM) to favor dynamic coarsening of graph. We also theoretically analyze the approximate equivalency of VI-GMM to weighted graph cut~\cite{dhillon2007weighted}. Finally, EI-GMM and VI-GMM can be alternately stacked into an end-to-end optimization network.

In summary, our main contributions are four folds: i) propose an end-to-end Gaussian-induced convolutional neural network for graph representation; ii) propose edge-induced GMM to encode variations of different subgraphs; iii) derive vertex-induced GMM to perform dynamic coarsening of graphs, which is an approximation to the weighted graph cut; iv) verify the effectiveness of our method and report state-of-the-art results on several graph datasets.

\begin{figure*}[t]
	\centering
	\includegraphics[width=0.83\textwidth]{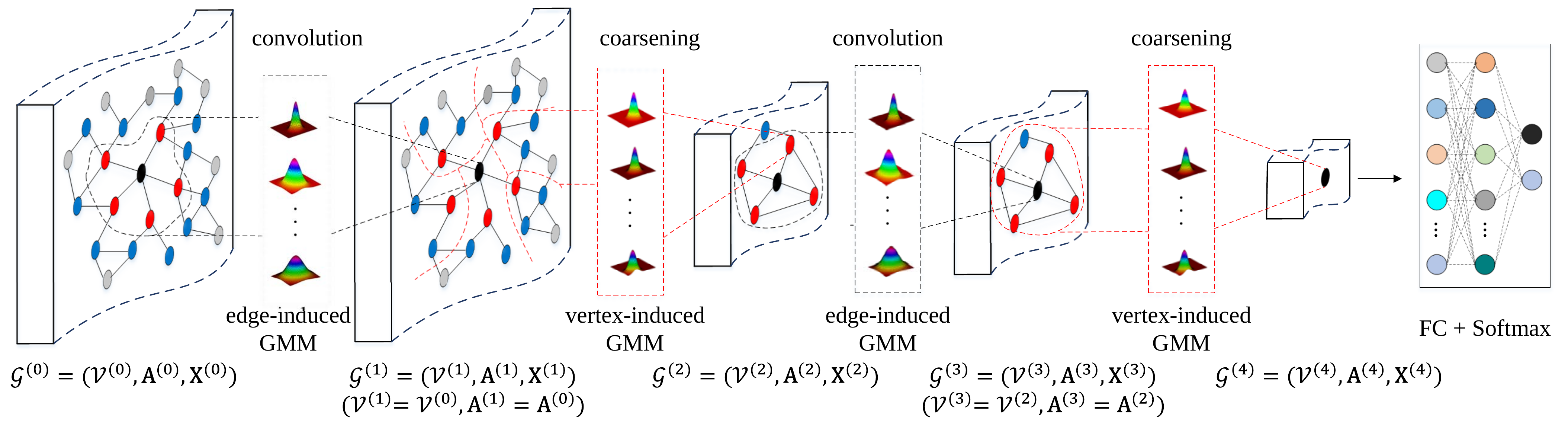}
	\caption{The GIC network architecture. The GIC main contains two module: convolution layer (EI-GMM) and coarsening layer (VI-GMM). The GIC stacks several convolution and coarsening layers alternatively and iteratively. More details can be found in incoming section.}
	\label{fig:network:A}
\end{figure*}

\section{Related Work}

Graph CNNs mainly fall into two categories: spectral and spatial methods. Spectral methods~\cite{bruna2013spectral,scarselli2009graph,henaff2015deep,such2017robust,li2018action,li2018spatio} construct a series of spectral filters by decomposing graph Laplacian, which often suffers high-computational burden. To address this problem, the fast local spectral filtering method~\cite{defferrard2016convolutional} parameterizes the frequency responses as a Chebyshev polynomial approximation. However, as shown in Fig.~\ref{fig:motivation:B}, after summarizing all nodes, this method will discard topology structures of a local receptive field. This kind of methods usually require equal sizes of graphs like the same sizes of images for CNNs~\cite{kipf2016semi}. Spatial methods attempt to define spatial structures of adjacent vertices and then perform filtering on structured graphs. Diffusion CNNs~\cite{atwood2016diffusion} scans a diffusion process across each node. PATCHY-SAN~\cite{niepert2016learning} linearizes neighbors by sorting weights of edges and deriving convolutional filtering on graphs, as shown in Fig.~\ref{fig:motivation:C}. As an alternative, random walks based approach is also used to define the neighborhoods~\cite{perozzi2014deepwalk}. For the linearized neighbors, RNNs~\cite{li2015gated} could be used to model the structured sequences. Similarly, NgramCNN~\cite{luo2017deep} serializes each graph by introducing the concept of $n$-gram block. 
GAT~\cite{velickovic2017graph} attempts to weight edges through the attention mechanism. WSC~\cite{jiang2018walk} attempts to aggregate walk fields defined by random walks into Gaussian mixture models. Zhao~\cite{zhao2018work} attempts to define a standard network with different graph convolutions. Besides, some variants~\cite{hamilton2017inductive,duran2017learning,zhang2018tensor} employ the aggregation or propagation of local neighbor nodes. Different from these tied filtering or ranking filtering methods, we use Gaussian models to encode local variations of graph. Also different from the recent mixture models~\cite{monti2017geometric}, which uses GMM to only learn the importance of adjacent nodes, our method uses weighted GMM to encode the distributions of local graph structures.

\section{The GIC Network}

\subsection{Attribute Graph}

Here we consider an undirected attribute graph $\mcG=(\mcV,\A,\X)$ of $m$ vertices (or nodes), where $\mcV=\{v_i\}_{i=1}^{m}$ is the set of vertices, $\A$ is a (weighted) adjacency matrix, and $\X$ is a matrix of graph attributes (or signals). The adjacency matrix $\A\in\mbR^{m\times m}$ records the connections between vertices. If $v_i, v_j$ are not connected, then $A(v_i,v_j)=0$, otherwise $A(v_i,v_j)\neq 0$. We sometimes abbreviate $A(v_i,v_j)$ as $A_{ij}$. The attribute matrix $\X\in\mbR^{m\times d}$ is associated with the vertex set $\mcV$, whose $i$-th row $\X_{i}$ (or $\X_{v_i}$) denotes a $d$-dimension attribute of the $i$-th node (\ie, $v_i$).

The graph Laplacian matrix $\L$ is defined as $\L = \D-\A$, where $\D\in\mbR^{m\times m}$ is the diagonal degree matrix with $D_{ii}=\sum_{j}A_{ij}$. The normalized version is written as $\L^{norm} = \D^{-1/2}\L\D^{-1/2}= \I-\D^{-1/2}\A\D^{-1/2}$.
where $\I$ is the identity matrix. Unless otherwise specified, we use the latter. We give the definition of subgraph used in the following.
\begin{defn} \label{def:graph}
	Given an attribute graph $\mcG=(\mcV,\A,\X)$, the attribute graph $\mcG'=(\mcV',\A',\X')$ is a subgraph of $\mcG$, denoted $\mcG'\subseteq\mcG$, if (i) $\mcV'\subseteq\mcV$, (ii) $\A'$ is the submatrix of $\A$ w.r.t. the subset $\mcV'$, and (iii) $\X'=\X_{\mcV'}$.
\end{defn}

\subsection{Overview}

The GIC network architecture is shown in Fig.~\ref{fig:network:A}. Given an attribute graph $\mcG^{(0)}=(\mcV^{(0)},\A^{(0)},\X^{(0)})$, where the superscript denotes the layer number, we construct multi-scale receptive fields for each vertex based on the adjacency matrix $\A^{(0)}$. Each receptive field records $k$-hop neighborhood relationships around the reference vertex, and forms a local centralized subgraph. To encode the centralized subgraph, we project it into edge-induced Gaussian models, each of which defines one variation ``direction" of the subgraph. We perform different filtering operations on different Gaussian components and aggregate all responses as the convolutional output. After the convolutional filtering, the input graph $\mcG^{(0)}$ is transformed into a new graph $\mcG^{(1)}=(\mcV^{(1)},\A^{(1)},\X^{(1)})$, where $\mcV^{(1)}=\mcV^{(0)}$ and $\A^{(1)}=\A^{(0)}$. To further abstract graphs, we next stack a coarsening layer on the graph $\mcG^{(1)}$. The proposed vertex-induced GMM is used to downsample the graph $\mcG^{(1)}$ into the low-resolution graph $\mcG^{(2)}=(\mcV^{(2)},\A^{(2)},\X^{(2)})$. Taking the convolution and coarsening modules, we may alternately stack them into a multi-layer GIC network, With the increase of layers, the receptive field size of filters will become larger, so the higher layer can extract more global graph information. In the supervised case of graph classification, we finally concatenate with a fully connected layer followed by a softmax loss layer.

\subsection{Multi-Scale Receptive Fields} 

In the standard CNN, receptive fields may be conveniently defined as latticed spatial regions. Thus convolution kernels on grid-shaped structures are accessible. However, the construction of convolutional kernels on graphs are intractable due to coordinate-free graphs, \eg, unordered vertices, unfixed number of adjacent edges/vertices. To address this problem, we resort to the adjacent matrix $\A$, which expresses connections between vertices. Since $\A^k$ exactly records the $k$-step reachable vertices, we may construct a $k$-neighbor receptive field by using the $k$-order polynomial of $\A$, denoted as $\psi_k(\A)$. Taking the simplest case, $\psi_k(\A)=\A^k$ reflects the $k$-hop neighborhood relationships. In order to remove the scale effect, we may normalize $\psi_k(\A)$ as $\psi_k(\A) \text{diag} (\psi_k(\A)\1)^{-1}$, which describes the reachable possibility in a $k$-hop walking. Formally, we define the $k$-th scale receptive field as a subgraph.
\begin{defn}\label{def:receptive}
	The $k$-th scale receptive field around a reference vertex $v_i$ is a subgraph $\mcG_{v_i}^k=(\mcV',\A',\X')$ of the k-order graph $(\mcV, \tbA=\psi_k(\A), \X)$, where $\mcV'=\{v_j|\wtA_{ij}\neq 0\}\cup\{v_i\}$, $\A'$ is the submatrix of $\tbA$ w.r.t. $\mcV'$, and $\X'=\X_{\mcV'}$.
\end{defn}

\subsection{Convolution: Edge-Induced GMM}

Given a reference vertex $v_i$, we can construct the centralized subgraph $\mcG_{v_i}^k$ of the $k$-th scale.
To coordinate the subgraph, we introduce Gaussian mixture models (GMMs), each of which may be understood as one principal direction of its variations. To encode the variations accurately, we jointly formulate attributes of vertices and connections of edges into Gaussian models. The edge weight $A'(v_i,v_j)$ indicates the relevance of $v_j$ to the central vertex $v_i$. The higher weight is, the stronger impact on $v_i$ is. So the weights can be incorporated into a Gaussian model by observing $A'(v_i,v_j)$ times. As the likelihood function, it is equivalent to raise the power $A'(v_i,v_j)$ on Gaussian function, which is proportional to $\mcN(\X'_{v_j}, \bmu, \frac{1}{A'(v_i,v_j)}\bSigma)$. Formally,  we estimate the probability density of the subgraph $\mcG_{v_i}^k$ from the $C_1$-component GMM,
\begin{align}
p_{v_i}(\X'_{v_j};\bTheta_1,A'_{ij}) &= \sum_{c=1}^{C_1}\pi_c\mcN(\X'_{v_j}; \bmu_c, \frac{1}{A'_{ij}}\bSigma_c), \nonumber \\
&\st \pi_c>0, \sum_{c=1}^{C_1}\pi_c=1, \label{eqn:GMM_edge}
\end{align}
where $\bTheta_1=\{\pi_1,\cdots,\pi_{C_1}, \bmu_1,\cdots,\bmu_{C_1},\bSigma_1,\cdots,\bSigma_{C_1}\}$ are the mixture parameters, $\{\pi_c\}$ are the mixture coefficients, $\{\bmu_c, \bSigma_c\}$ are the parameters of the $k$-th component, and $A'_{ij}>0$ \footnote{In practice, we normalize $\A'$ into a non-negative matrix.}. Intuitively, edge weight $A'_{ij}$ is, the stronger impact of the node $v_j$ w.r.t. the reference vertex $v_i$ is. We will refer to the model in Eqn.~(\ref{eqn:GMM_edge}) as the \textit{edge-induced Gaussian mixture model} (EI-GMM).

In what follows, we assume all attributes of nodes are independent on each other, which is often used in signal processing. That means, the covariance matrix $\bSigma_c$ is diagonal, so we denote it as $\text{diag}(\bsigma_c^2)$. To avoid the explicit constraints for $\pi_c$ in Eqn.~(\ref{eqn:GMM_edge}), we adopt the soft-max normalization with the re-parameterization variable $\alpha_c$, \ie, $\pi_c = {\exp(\alpha_c)}/{\sum_{k=1}^{C_1}\exp(\alpha_k)}$. Thus, the entire subgraph log-likelihood can be written as
\begin{align}
\zeta(\mcG_{v_i}^k) &= \sum_{j=1}^m\ln p_{v_i}(\X'_{v_j};\bTheta_1,\A') \nonumber \\
&= \sum_{j=1}^m\ln\sum_{c=1}^{C_1}\pi_c\mcN(\X'_{v_j}; \bmu_c, \frac{1}{A'_{ij}}\bSigma_c),\label{eqn:GMM_log}
\end{align}
To infer forward, instead of the expectation-maximization (EM) algorithm, we use the gradients of the subgraph with regard to the parameters of the EI-GMM model $\bTheta_1$, motivated by the recent Fisher vector work~\cite{sanchez2013image}, which has been proven to be effective in representation.

For a convenient calculation, we simplify the notations, $\mcN_{jc} = \mcN(\X'_{v_j}, \bmu_c, \frac{1}{A'_{ij}}\bsigma_c^2)$ and $ Q_{jc}=\frac{\pi_c\mcN_{jc} }{\sum_{k=1}^{C_1}\pi_k\mcN_{jk}}$, then we can derive the gradients of model parameters from Eqn.~(\ref{eqn:GMM_log}) as follows
\begin{align}
&\frac{\partial \zeta(\mcG_{v_i}^k)}{\partial\bmu_c}  \!\!=\!\! \sum_{j=1}^m\frac{A'_{ij}Q_{jc}(\X'_{v_j}-\bmu_c)}{\bsigma_c^2}, \quad \nonumber \\
&\frac{\partial \zeta(\mcG_{v_i}^k)}{\partial\bsigma_c}  \!\!=\!\! \sum_{j=1}^m \frac{Q_{jc}(A'_{ij}(\X'_{v_j}-\bmu_c)^2-\bsigma_c^2)}{\bsigma_c^3},
\end{align} 
where the division of vectors means a term-by-term operation. Note we do not use $\partial \zeta(\mcG_{v_i}^k) / \partial\alpha_c$ due to no improvement in our experience. The gradients describe the contribution of the corresponding parameters to the generative process. The subgraph variations are adaptively allocated to $C_1$ Gaussian models. Finally, we ensemble all gradients w.r.t. Gaussian model (\ie, directions of graph) to analogize the collection of local square receptive field on image. Formally, for the $k$-scale receptive field $\mcG_{v_i}^k$ around the vertex $v_i$, the attributes produced from Gaussian models are filtered respectively and then concatenated,
\begin{align}
F(\mcG_{v_i}^k,\bTheta_1,f) &= \text{ReLU}(\sum_{c=1}^{C_1} f_i(\text{Cat}[\frac{\partial \zeta(\mcG_{v_i}^k)}{\partial\bmu_c},\frac{\partial \zeta(\mcG_{v_i}^k)}{\partial\bsigma_c}]), \label{eqn:gau_fea}
\end{align}
where $\text{Cat}[\cdot, \cdot]$ is a concatenation operator, $f_i$ is a linear filtering function (\ie, a convolution function) and ReLU is the rectified linear unit. Therefore we can produce the feature vectors that have same dimensionality depending on the number of Gaussian models for different subgraphs. If the soft assignment distribution $ Q_{jc} $ is sharply peaked on a single value of one certain Gaussian for the vertex $v_{j}$, the vertex will be only projected onto one Gaussian direction.

\subsection{Coarsening: Vertex-Induced GMM}

Like the standard pooling in CNNs, we need to downsample graphs so as to abstract them as well as reduce the computational cost. However, the pooling on images are tailored for latticed structures, and cannot be used for irregular graphs. One solution is to use some clustering algorithms to partition vertices to several clusters, and then produce a new vertex from each cluster. However, we expect that two vertices should not fall into the same cluster with a larger possibility if there is a high transfer difficulty between them. To this end, we derive vertex-induced Gaussian mixture models (VI-GMM) to weight each vertex. To utilize the edge information,  we construct a latent observation $\phi(v_i)$ w.r.t. each vertex $v_i$ from the graph Laplacian (or adjacent matrix if semi-positive definite), \ie, the kernel calculation $\langle\phi(v_i),\phi(v_j)\rangle=L_{ij}$. Moreover, for each vertex $v_i$, we define an influence factor $w_i$ for Gaussian models. Formally, given $C_2$ Gaussian models, VI-GMM is written as
\begin{align}
p(\phi(v_i);\bTheta_2,w_i) &= \sum_{c=1}^{C_2}\pi_c\mcN(\phi(v_i); \bmu_c, \frac{1}{w_i}\bSigma_c), \nonumber \\
&\st  w_i=h(\X_{v_i})>0,
\end{align}
where $h$ is a mapping function to be learnt. To reduce the computation cost of matrix inverse on $\bSigma$, we specify it as an identity matrix. Then we have
\begin{align}
p(\phi(v_i);\bTheta_2,w_i) =  \sum_{c=1}^{C_2}\frac{\pi_c}{(\frac{2\pi}{w_i})^{d/2}}\exp^{-\frac{w_i}{2}\|\phi(v_i)-\bmu_c\|^2}, 
\end{align}
Given a graph with $m$ vertices, the objective is to maximize the following log-likelihood:
\begin{align}
\argmax_{\bTheta_2} \zeta(\bTheta_2) = \sum_{i=1}^m \ln\sum_{c=1}^{C_2}\pi_c\mcN(\phi(v_i); \bmu_c, \frac{1}{w_i}\I)). \label{eqn:likelihood}
\end{align}

To solve above model in Eqn.~(\ref{eqn:likelihood}), we use the iterative expectation maximization algorithm, which has closed-form solution at each step. Meanwhile, the algorithm may automatically conduct the required constraints. The graphical clustering process is summarized as follows:

(1) {E-Step}: the posteriors, \ie, the $i$-th vertex for the $c$-th cluster, are updated with
$p_{ic} = \frac{\pi_c p(\phi(v_i);\btheta_c, w_i)}{\sum_{k=1}^C \pi_k p(\phi(v_i);\btheta_k, w_i)}$,
where $\btheta_c$ is the $c$-th Gaussian parameters, and $\bTheta_2=\{\btheta_1,\cdots,\btheta_{C_2}\}$.

(2) {M-Step}: we optimize Gaussian parameters $\pi, \bmu$. The parameter estimatation is given by $
\pi_c = \frac{1}{m}\sum_{i=1}^m r_{ic},
\bmu_c =\frac{\sum_{v_i\in \mcG_c}w_i\phi(v_i)}{\sum_{v_i\in \mcG_c}w_i}$.
$\pi_c$ indicates the energy summation of all vertices assigned to the cluster $c$, and $\bmu_c$ may be understood as a doubly weighted ($w_i, r_{ic}$) average on the cluster $c$.

After several iterations of the two steps, we perform hard quantification. The $i$-th vertex is assigned as the class with the maximum possibility, formally, $r_{ic} = 1$ if $c=\argmax_{k} p_{ik}$, otherwise 0. Thus we can obtain the cluster matrix $\P\in\{0,1\}^{m\times C_2}$, where $P_{ic}=1$ if the $i$-th vertex falls into the cluster $c$. During coarsening, we take maximal responses of each cluster as the attributes of new vertex, and derive a new adjacency matrix by using $\P\tp\A\P$.

It is worth noting that we need not compute the concrete $\phi$ during the clustering process. The main calculation $\|\phi(v_i)-\bmu_c\|^2$ in EM can be reduced to the kernel version:
$K_{ii}-\frac{2\sum_{v_j\in\mcG_c}w_jK_{ij}}{\sum_{v_j\in\mcG_c}w_j}
+\frac{\sum_{v_j,v_k\in\mcG_c}w_jw_k K_{jk}}{(\sum_{v_j\in\mcG_c}w_j)^2}$,
where $K_{ij} = \langle\phi(v_i),\phi(v_j)\rangle$.
In practice, we can use the graph Laplacian $\L$ as the kernel. In this case, we can easily reach the following proposition, which is relevant to graph cut~\cite{dhillon2007weighted}.
\begin{prop}
	In EM, if the kernel matrix takes the weight-regularized graph Laplacian, \ie, $\mcK= \text{diag}(\w)\L \text{diag}(\w)$, then VI-GMM is equal to an approximate optimization of graph cut, \ie, $
	\min \sum_{c=1}^C\frac{\text{links}(\mcV_c, \mcV\backslash \mcV_c)}{w(\mcV_c)}$,
	where $\text{links}(\mcA, \mcB)=\sum_{v_i\in\mcA,v_j\in\mcB} A_{ij}$, and $w(\mcV_c)=\sum_{j\in\mcV_c}w_j$.
\end{prop}

\section{Experiments}

\subsection{Graph Classification}

For graph classification, each graph is annotated with one label. We use two types of datasets: Bioinformatics and Network datasets. The former contains MUTAG~\cite{debnath1991structure}, PTC~\cite{toivonen2003statistical}, NCI1 and NCI109~\cite{wale2008comparison}, ENZYMES~\cite{borgwardt2005protein} and PROTEINS~\cite{borgwardt2005protein}. The latter has COLLAB~\cite{leskovec2005graphs}, REDDIT-BINARY, REDDIT-MULTI-5K, REDDIT-MULTI-12K, IMDB-BINARY and IMDB-MULTI.

\begin{table*}[!t]
	\centering
	\caption{Comparisons with state-of-the-art methods.}
	\begin{sc}
	\scalebox{0.85}{
		\begin{tabular}{|l| c c c| c c | c c c |c |c |c |c|}
			\toprule
			Dataset 
			&PSCN  &DCNN  &NgramCNN   &FB  &DyF   &WL   &GK  &DGK  &RW   &SAEN  & GIC \\
			
			\midrule
			\multirow{2}{*}{MUTAG}
			&92.63   &66.98  &\textbf{94.99}  &84.66   &88.00    &78.3   &81.66     &82.66      &83.72     &84.99 &94.44   \\		
			& $\pm$4.21  &--    &\textbf{$\pm$5.63} &$\pm$2.01 & $\pm$2.37  & $\pm$1.9 & $\pm$2.11  & $\pm$1.45  & $\pm$1.50   & $\pm$1.82  	&$\pm$4.30      \\
			
			\midrule
			\multirow{2}{*}{PTC}
			&60.00   &56.60  &68.57  &55.58   &57.15     &--     &57.26     &57.32   &57.85    &57.04  &\textbf{77.64}   \\			
			& $\pm$4.82  &--   &$\pm$1.72  &2.30   & $\pm$1.47  & --  &  $\pm$1.41  &  $\pm$1.13  & $\pm$1.30
			&  $\pm$ 1.30  & \textbf{$\pm$ 6.98}      \\
			
			\midrule
			\multirow{2}{*}{NCI1}
			&78.59   &62.61     &--  &62.90   &68.27  &83.1   &62.28   &62.48   &48.15    &77.80  &\textbf{84.08}    \\			
			& $\pm$1.89  &--   &--  &$\pm$0.96  & $\pm$0.34  & $\pm$0.2  & $\pm$0.29  &  $\pm$0.25  &  $\pm$0.50  & $\pm$ 0.42  & \textbf{$\pm$1.77}     \\
			
			\midrule
			\multirow{2}{*}{NCI109}
			& --   &62.86    &--  &62.43  & 66.72   & \textbf{85.2}    & 62.60   & 62.69   & 49.75 	 & --  & 82.86    \\			
			& --   &--   &--  &$\pm$1.13    & $\pm$ 0.20  & \textbf{$\pm$ 0.2}  & $\pm$ 0.19  &  $\pm$ 0.23  &  $\pm$ 0.60	& --  & $\pm$ 2.37     \\
			
			\midrule
			\multirow{2}{*}{ENZYMES}
			& --    &18.10    &--  &29.00   & 33.21   & 53.4   & 26.61   & 27.08   & 24.16   & --   & \textbf{62.50}  \\			
			& --    &--    &--  &$\pm$1.16  & $\pm$ 1.20  & $\pm$ 1.4   &  $\pm$ 0.99  & $\pm$ 0.79  &  $\pm$ 1.64
			&--  & \textbf{$\pm$ 5.12}     \\
			
			\midrule
			\multirow{2}{*}{PROTEINS}
			& 75.89  &--   &75.96  &69.97 & 75.04   & 73.7  & 71.67   & 71.68  & 74.22    & 75.31  & \textbf{77.65}     \\			
			& $\pm$ 2.76  &--   &$\pm$2.98  &$\pm$1.34  & $\pm$ 0.65  &  $\pm$ 0.5  & $\pm$ 0.55  &  $\pm$ 0.50  & $\pm$ 0.42	&  $\pm$ 0.70 & \textbf{$\pm$ 3.21}      \\
			
			\midrule
			\multirow{2}{*}{COLLAB}
			& 72.60  &--    &--  &76.35 & 80.61 & -- & 72.84 & 73.09 & 69.01 & 75.63 & \textbf{81.24}     \\			
			& $\pm$ 2.15  &--   &--  &1.64  & $\pm$ 1.60  & --  & $\pm$ 0.28  & $\pm$ 0.25  & $\pm$ 0.09
			& $\pm$ 0.31  & \textbf{$\pm$ 1.44}      \\
			
			\midrule
			\multirow{2}{*}{REDDIT-B}
			& 86.30  &--     &--  &88.98 &\textbf{89.51}  &75.3 & 77.34 & 78.04 & 67.63  & 86.08 & 88.45   \\			
			& $\pm$ 1.58  &--   &--  &$\pm$2.26  & \textbf{$\pm$ 1.96}  & $\pm$ 0.3  & $\pm$ 0.18  & $\pm$ 0.39  & $\pm$ 1.01
			& $\pm$ 0.53  & $\pm$ 1.60     \\
			
			\midrule
			\multirow{2}{*}{REDDIT-5K}
			& 49.10  &--      &--  &50.83   & 50.31 & --  & 41.01 & 41.27 & --  &\textbf{52.24} &51.58 \\		
			& $\pm$ 0.70  &--    &--  &1.83  &$\pm$ 1.92  & --  & $\pm$ 0.17  & $\pm$ 0.18  & --
			& \textbf{$\pm$ 0.38}  & $\pm$ 1.68   \\
			
			\midrule
			\multirow{2}{*}{REDDIT-12K}
			& 41.32  &--      &--  &42.37  & 40.30 & --  & 31.82 & 32.22 & --  & \textbf{46.72} & 42.98 \\			
			& $\pm$ 0.42  &--    &--  &1.27   & $\pm$ 1.41  & -- & $\pm$ 0.08  & $\pm$ 0.10  & --
			& \textbf{$\pm$ 0.23} & $\pm$ 0.87     \\
			
			\midrule
			\multirow{2}{*}{IMDB-B}
			& 71.00   &--    &71.66  &72.02 & 72.87 & 72.4 & 65.87 & 66.96 & 64.54 	& 71.26 & \textbf{76.70}     \\			
			& $\pm$ 2.29  &--    &$\pm$2.71  &$\pm$4.71  & $\pm$ 4.05  & $\pm$ 0.5  & $\pm$ 0.98  & $\pm$ 0.56  & $\pm$ 1.22
			& $\pm$ 0.74  & \textbf{$\pm$ 3.25}      \\
			
			\midrule
			\multirow{2}{*}{IMDB-M}
			& 45.23 &--     &50.66  &47.34  & 48.12 & -- & 43.89 & 44.55 & 34.54 & 49.11 & \textbf{51.66}      \\			
			& $\pm$ 2.84  &--  &$\pm$4.10  &3.56  & $\pm$ 3.56  & -- & $\pm$ 0.38  & $\pm$ 0.52  & $\pm$ 0.76
			& $\pm$ 0.64  & \textbf{$\pm$ 3.40}      \\
			\bottomrule
		\end{tabular}
	}
	\end{sc}
	\label{table:state-of-the-art}
\end{table*}
\begin{table}[!t]
	\centering
	\caption{Node label prediction on Reddit and PPI data (micro-averaged F1 score).}
	\label{table:multi-label}
	\begin{sc}	
		\scalebox{0.9}{
			\begin{tabular}{l c c}
				\toprule
				Dataset      & Reddit  & PPI   \\
				\midrule	
				Random       & 0.042   & 0.396                     \\
				Raw features & 0.585   & 0.422                       \\
				Deep walk    & 0.324   & --                       \\
				Deep walk + features  & 0.691  & --               \\
				Node2Vec + regression  & 0.934 & -- \\
				GraphSAGE-GCN  & 0.930   & 0.500                       \\			
				GraphSAGE-mean  & 0.950   & 0.598                       \\
				GraphSAGE-LSTM  & \textbf{0.954}   & 0.612                       \\
				GIC    & 0.952   & \textbf{0.661}                     \\
				\bottomrule
			\end{tabular}
		}
	\end{sc}
\end{table}

\subsubsection{Experiment Settings} 

We verify our GIC on the above bioinformatics and social network datasets. In default, GIC mainly consists of three graph convolution layers, each of which is followed by a graph coarsening layer, and one fully connected layer with a final softmax layer as shown in Fig~\ref{fig:network:A}. Its configuration can simply be set as C(64)-P(0.25)-C(128)-P(0.25)-C(256)-P-FC(256), where C, P and FC denote the convolution, coarsening and fully connected layers respectively. The choices of hyperparameters are mainly inspired from the classic VGG net. For example, the coarsening factor is 0.25 (w.r.t. 0.5$\times$0.5 in VGG), the attribute dimensions at three conv. layers are 64-128-256 (w.r.t. the channel numbers of conv1-3 in VGG). The scale of respective field and the number of Gaussian components are both set to 7. We train GIC network with stochastic gradient descent for roughly 300 epochs with a batch size of 100, where the learning rate is 0.1 and the momentum is 0.95. 

In the bioinformatics datasets, we exploit labels and degrees of the vertices to generate initial attributes of each vertex. In the social network datasets, we use degrees of vertices. We closely follow the experimental setup in PSCN~\cite{niepert2016learning}. We perform 10-fold cross-validation, 9-fold for training and 1-fold for testing. The experiments are repeated 10 times and the average accuracies are reported.

\subsubsection{Comparisons with the State-of-the-arts}

We compare our GIC with several state-of-the-arts, which contain graph convolution networks (PSCN~\cite{niepert2016learning}, DCNN~\cite{atwood2016diffusion}, NgramCNN~\cite{luo2017deep}), neural networks (SAEN~\cite{orsini2017shift}), feature based algorithms (DyF~\cite{gomez2017dynamics}, FB~\cite{bruna2013spectral}), random walks based methods (RW~\cite{gartner2003graph}), graph kernel approaches (GK~\cite{shervashidze2009efficient}, DGK~\cite{yanardag2015deep}, WL~\cite{morris2017glocalized}). We present the comparisons with the state-of-the-arts, as shown in Table~\ref{table:state-of-the-art}. All results come from the related literatures. We have the following observations.

Deep learning based methods on graphs (including DCNN, PSCN, NgramCNN, SAEN and ours) are superior to those conventional methods in most cases. The conventional kernel methods usually require the calculation on graph kernels with high-computational complexity. In contrast, these graph neural networks attempt to learn more abstract high-level features by performing inference-forward, which need relatively low computation cost.

Compared with recent graph convolution methods, ours can achieve better performance on most datasets, such as PTC, NCI1, NCI109, ENZYMES and PROTEINS. The main reason should be that local variations of subgraphs are accurately described with Gaussian component analysis.

The proposed GIC achieves state-of-the-art results on most datasets. The best performance is gained in some bioinformatics datasets and some social network datasets including PTC, NCI1, ENZYMES, PROTEINS, COLLAB, IMDB-BINARY and IMDB-MULTI. Although NgramCNN, DyF, WL and SEAN approaches have obtained the best performance on MUTAG, REDDIT-BINARY, NCI109, REDDIT-MULTI-5K and REDDIT-MULTI-12K respectively, our method is fully comparable to them.

\subsection{Node Classification}

For node classification, one node is assigned one/multiple labels. It is challenging if the label set is large. During training, we only use a fraction of nodes and their labels. The task is to predict the labels for the remaining nodes. Following the setting in~\cite{hamilton2017inductive}, we conduct the experiments on Reddit data and PPI data. For a fair comparison to graphSAGE~\cite{hamilton2017inductive}, we use the same initial graph data, mini-batch iterators, supervised loss function and neighborhood sample. The other network parameters are similar to graph classification except removing the coarsening layer.

Tabel~\ref{table:multi-label} summarizes the comparison results. Our GIC can obtain the best performance 0.661 on PPI data and a comparable result 0.952 on Reddit data. The raw features provide an important initial information for node multi-label classification. Based on the raw features, deep walk~\cite{perozzi2014deepwalk} improves about 0.36 (micro-F1 scores) on Reddit data. Meanwhile, we conduct an experiment of node2vec and use regression model to classification. Our method gains better performance than node2vec~\cite{grover2016node2vec}. Comparing different aggregation methods like GCN~\cite{kipf2016semi}, mean and LSTM, our GIC has a significant improvement about 0.16 on PPI data and gains a competitive performance on Reddit data. The results demonstrate our approach is robust to infer unknown labels of partial graphs.

\subsection{Model Analysis}

\begin{table}[!t]
	\centering
	\caption{The verification of our convolution and coarsening.}
	\label{table:convandpooling}
	\begin{sc}
		\scalebox{0.68}{
			\begin{tabular}{l c c c c}
				\toprule
				\multirow{2}{*}{Dataset}        & ChebNet   & GCN   & GIC  &\multirow{2}{*}{GIC} \\
				& w/ VI-GMM   & w/ VI-GMM & w/o VI-GMM & \\
				\midrule
				MUTAG          & 89.44 $\pm$ 6.30      & 92.22 $\pm$ 5.66   & 93.33 $\pm$ 4.84  & \textbf{94.44 $\pm$ 4.30} \\
				PTC            & 68.23 $\pm$ 6.28      & 71.47 $\pm$ 4.75   & 68.23 $\pm$ 4.11  & \textbf{77.64 $\pm$ 6.98} \\
				NCI1           & 73.96 $\pm$ 1.87      & 76.39 $\pm$ 1.08   & 79.17 $\pm$ 1.63  & \textbf{84.08 $\pm$ 1.77}	\\		
				NCI109         & 72.88 $\pm$ 1.85      & 74.92 $\pm$ 1.70   & 77.81 $\pm$ 1.88  & \textbf{82.86 $\pm$ 2.37} \\
				ENZYMES        & 52.83 $\pm$ 7.34      & 51.50 $\pm$ 5.50   & 52.00 $\pm$ 4.76  & \textbf{62.50 $\pm$ 5.12} \\
				PROTEINS       & 78.10 $\pm$ 3.37      & \textbf{80.09 $\pm$ 3.20}    & 78.19 $\pm$ 2.04 & 77.65 $\pm$ 3.21 \\
				\bottomrule
			\end{tabular}
		}
	\end{sc}
\end{table}
\begin{table}[!t]
	\centering
	\caption{Comparisons on $K$ and $C_1$.}
	\label{table:GMM}
	\begin{sc}
		\scalebox{0.68}{
			\begin{tabular}{l c c c c}
				\toprule
				Dataset        & $K,C_1=1$                & $K,C_1=3$                  & $K,C_1=5$                  & $K,C_1=7$       \\
				\midrule
				MUTAG         & 67.77 $\pm$ 11.05        & 83.88 $\pm$ 5.80           & 90.55 $\pm$ 6.11           & \textbf{94.44 $\pm$ 4.30}    \\
				PTC           & 72.05 $\pm$ 8.02         & 77.05 $\pm$ 4.11           & 76.47 $\pm$ 5.58           & \textbf{77.64 $\pm$ 6.98}    \\
				NCI1         & 71.21 $\pm$ 1.94         & 83.26 $\pm$ 1.17        & \textbf{84.47 $\pm$ 1.64}  & 84.08 $\pm$1.77\\				
				NCI109         & 70.02 $\pm$ 1.57         & 81.74 $\pm$ 1.56           & \textbf{83.39 $\pm$ 1.65}  & 82.86 $\pm$ 2.37            \\
				ENZYMES        & 33.83 $\pm$ 4.21         & \textbf{64.00 $\pm$ 4.42}  & 63.66 $\pm$ 3.85           & 62.50 $\pm$ 5.12            \\
				PROTEINS       & 75.49 $\pm$ 4.00         & 77.47 $\pm$ 3.37           & \textbf{78.10 $\pm$ 2.96}  & 77.65 $\pm$ 3.21            \\
				\bottomrule			
				
			\end{tabular}
		}
	\end{sc}
\end{table}
\begin{table}[!t]
	\centering
	\caption{Comparisons on the layer number.}
	\label{table:layer}
	\begin{sc}	
		\scalebox{0.68}{
			\begin{tabular}{l c c c c}
				\toprule
				Dataset       & $N=2$                  & $N=4$                        & $N=6$                      & $N=8$          \\
				\midrule
				MUTAG         & 86.66 $\pm$ 8.31       & 91.11 $\pm$ 5.09             & 93.88 $\pm$ 5.80           & \textbf{94.44 $\pm$ 4.30}   \\
				PTC           & 64.11 $\pm$ 6.55       & 74.41 $\pm$ 6.45             & 75.29 $\pm$ 6.05           & \textbf{77.64 $\pm$ 6.98}   \\
				NCI1          & 71.82 $\pm$ 1.85       & 81.36 $\pm$ 1.07             & 83.01 $\pm$ 1.54           & \textbf{84.08 $\pm$ 1.77}   \\
				NCI109        & 71.09 $\pm$ 2.41       & 80.02 $\pm$ 1.67             & 81.60 $\pm$ 1.83           & \textbf{82.86 $\pm$ 2.37}    \\
				ENZYMES       & 42.33 $\pm$ 4.22       & 61.83 $\pm$ 5.55             & \textbf{64.83 $\pm$ 6.43}  & 62.50 $\pm$ 5.12          \\
				PROTEINS      & 77.38 $\pm$ 2.97       & \textbf{79.81 $\pm$ 3.84}    & 78.37 $\pm$ 4.00           & 77.65 $\pm$ 3.21          \\
				\bottomrule
			\end{tabular}
		}
	\end{sc}
\end{table}

\textbf{EI-GMM and VI-GMM}: To directly analyze convolution filtering with EI-GMM, we compare our method with ChebNet~\cite{defferrard2016convolutional} and GCN~\cite{kipf2016semi} approaches by using the same coarsening mechanism VI-GMM.  As shown in Table~\ref{table:convandpooling}, under the same coarsening operation, our GIC is superior to ChebNet+VI-GMM and GCN+VI-GMM. It indicates EI-GMM can indeed encode the variations of subgraphs more effectively. On the other hand, we remove the coarsening layer from our GIC. For different size graphs, we pad new zero vertices into a fixed size and then concatenate attributes of all vertices for classification. As shown in this table, the performance of GIC still outperforms GIC without VI-GMM coarsening, which verifies the effectiveness of the coarsening layer VI-GMM.

\textbf{$K$ and $C_1$}: The kernel size $K$ and the number of Gaussian components $C_1$ are the most crucial parameters. Generally, the $C_1$ is proportional to the $K$. The reason is that the larger receptive field usually contains more vertices (\ie, a relative large subgraph). Thus we simply take the equal values for them, $K=C_1 = \{1,3,5,7\}$. The experimental results are shown in Table~\ref{table:GMM}. With the increase of $K, C_1$, the performance improves at most cases. The reasons are two folds: i) with increasing receptive field size, the convolution will cover the farther hopping neighbors; ii) with the increase of $C_1$, the variations of subgraphs are encoded more accurately. But for the larger values of $K$ and $C_1$ will increase the computational burden. Moreover, the overfitting phenomenon might occur with the increase of model complexity. Take the example of NCI109, in the first convolution layer, the encoded attributes (in Eqn.~(\ref{eqn:gau_fea})) will be $2\times39\times7=546$ for each scale of receptive field, where $39$ is the dimension of attributes (w.r.t the number of node labels) and $7$ is the number of Gaussian components. Thus, for 7 scales of receptive field, the final encoded attributes will be $546\times7=3822$ dimensions, which will be mapping to 64 dimensions by the function $f=[f_1,\cdots,f_{C_1}]$ in Eqn.~(\ref{eqn:gau_fea}). Thus the model parameter is $3822\times64=244608$ in the first layer. Similarly, if the number of node label is 2, the model parameter will sharply decrease into $18816$. Besides, the parameter complexity is related to the number of classes and nodes. The comparison results in Table~\ref{table:GMM} demonstrate the trend of the parameters $K$ and $C_1$ in our GIC framework.

\textbf{Number of stacked layers}: Here we test on the number of stacked network layers with $N=2,4,6,8$. When $N=2$, only one fully connected layer and one softmax layer are preserved. When $N=4$, we add two layers: the convolution layer and the coarsening layer. When continuing to stack both, the depth of network will be 6 and 8. The results are shown in Table~\ref{table:layer}. Deeper networks can gain better performance in most cases, because the larger receptive field is observed and more abstract structures will be extracted in the topper layer. Of course, there is an extra risk of overfitting due to the increase of model complexity.

\textbf{An analysis of computation complexity}: In the convolution layer, the computational costs of receptive fields and Gaussian encoding are about $O(Km^2)$ and $O(C_1d^2)$ respectively, where $m, d$ are number of nodes and the feature dimensionality. Generally, $K=C_1\ll d<m$. In the coarsening layer, the time complexity is about $O(pm^2+md)$, where $p$ is iteration number of the EM algorithm. In all, suppose the whole GIC alternatively stacks $n$ convolution and coarsening layers, the entire time complexity is $O(n(K+p)m^2+nC_1d^2+nmd)$.

\section{Conclusion}
In this paper, we proposed a novel Gaussian-induced convolution network to handle with general irregular graph data. Considering the previous spectral and spatial methods do not well characterize local variations of graph, we derived edge-induced GMM to adaptively encode subgraph structures by projecting them into several Gaussian components and then performing different filtering operations on each Gaussian direction like the standard CNN filters on images. Meanwhile, we formulated graph coarsening into vertex-induced GMM to dynamically partition a graph, which was also proven to be equal to graph cut. Extensive experiments in two graphic tasks (i.e. graph and node classification) demonstrated the effectiveness and superiority of our GIC compared with those baselines and state-of-the-art methods. In the future, we would like to extend our method into more applications to irregular data.

\section{Acknowledgments}
The authors would like to thank the Chairs and the anonymous reviewers for their critical and constructive comments and suggestions. This work was supported by the National Science Fund of China under Grant Nos. 61602244, 61772276, U1713208 and 61472187 and Program for Changjiang Scholars.

\small
\bibliographystyle{aaai}
\bibliography{ref}

\end{document}